\documentclass[10pt,twocolumn,letterpaper]{article}

\usepackage{cvpr} 
\usepackage{times}
\usepackage{epsfig}
\usepackage{stfloats}
\usepackage{graphicx}
\usepackage{amsmath}
\usepackage{amsthm}
\usepackage{amssymb}
\usepackage{algorithm}
\usepackage{tikz}
\usepackage{algpseudocode}

\usepackage{float}
\usepackage{subcaption} 
\usepackage{booktabs}
\usepackage{multirow}
\usepackage{bm}

\usepackage{geometry} 
\usepackage{amsmath,graphicx}
\usepackage{enumitem}

\usetikzlibrary{shapes, arrows, shapes.misc, positioning}
\usetikzlibrary{shapes.geometric, arrows.meta, positioning, shadows}
\usetikzlibrary{arrows.meta, positioning, shapes.symbols}
\usetikzlibrary{shapes.geometric, arrows.meta, positioning}

\usepackage[pagebackref=true,breaklinks=true,letterpaper=true,colorlinks,bookmarks=false]{hyperref}

\begin{document}

\title{CURE: Centroid-guided Unsupervised Representation Erasure for Facial Recognition Systems}

\author{
FNU Shivam \quad
Nima Najafzadeh \quad
Yenumula Reddy \quad
Prashnna Gyawali \\
West Virginia University \\
Morgantown, USA \\
{\tt\small \{ss00132, nn00008\}@mix.wvu.edu, \{ramana.reddy, prashnna.gyawali\}@mail.wvu.edu}
}


\maketitle
\thispagestyle{empty}

\begin{abstract}

In the current digital era, facial recognition systems offer significant utility and have been widely integrated into modern technological infrastructures; however, their widespread use has also raised serious privacy concerns, prompting regulations that mandate data removal upon request. Machine unlearning has emerged as a powerful solution to address this issue by selectively removing the influence of specific user data from trained models while preserving overall model performance. However, existing machine unlearning techniques largely depend on supervised techniques requiring identity labels, which are often unavailable in privacy-constrained situations or in large-scale, noisy datasets. To address this critical gap, we introduce CURE (Centroid-guided Unsupervised Representation Erasure), the first unsupervised unlearning framework for facial recognition systems that operates without the use of identity labels, effectively removing targeted samples while preserving overall performance. We also propose a novel metric, the Unlearning Efficiency Score (UES), which balances forgetting and retention stability, addressing shortcomings in the current evaluation metrics. CURE significantly outperforms unsupervised variants of existing unlearning methods. Additionally, we conducted quality-aware unlearning by designating low-quality images as the forget set, demonstrating its usability and benefits, and highlighting the role of image quality in machine unlearning. The full code can be found here: \url{https://github.com/Shivam101s/CURE_FaceUnlearning}

\end{abstract}

\section{Introduction}

Modern facial recognition systems have become a regular part of today's technology. They are everywhere in modern society, helping build our digital infrastructure, making verification and identification easier, and improving surveillance. These systems use deep learning methods to achieve high accuracy in real-world applications \cite{wang2021deep}. However, the growing use of facial recognition has raised major concerns about privacy, data protection, and potential misuse. These issues are well-discussed in research, showing the ethical challenges that come with large-scale use \cite{almeida2022ethics}. This has created a clear tension between protecting privacy and ensuring public safety.

Recently, legal provisions like the European Union’s General Data Protection Regulation (GDPR) \cite{book} and the California Consumer Privacy Act (CCPA) \cite{goldman2020introduction} have added the ``right to be forgotten", requiring organizations to delete an individual's data from trained models upon request. These requirements have pushed the development of a new paradigm of \textit{Machine Unlearning}, introduced as a way to meet privacy needs \cite{cao2015towards}. It focuses on removing the effect of certain data points (\textit{forget set}) from trained models while keeping the model’s performance on the remaining data (\textit{retain set}).

Despite its importance, the problem of unlearning in facial recognition systems is still not well studied. Most of the existing work on machine unlearning—such as SCRUB \cite{kurmanji2023towards}, Negative Gradient \cite{golatkar2020eternal}, and Bad Teacher \cite{chundawat2023can}—has been developed in general machine learning contexts, not specifically for facial recognition. These methods are mostly supervised and rely on ground-truth labels for both forget and retain samples, which is a major limitation. In real-world facial recognition scenarios, getting such labels may not be possible \cite{bourtoule2021machine}, especially due to privacy concerns or when working with large-scale, noisy, incomplete, or legacy datasets. This makes existing approaches less practical for real-world, privacy-constrained facial recognition systems.

In this work, we pioneer the exploration of unsuprvised unlearning within facial recognition system by addressing the challenge of unlearning in facial recognition systems without  relying on identity labels. 
We proposed the first unsupervised unlearning framework, called \textbf{CURE (Centroid-guided Unsupervised Representation Erasure)}.
CURE introduces a centroid-guided contrastive unlearning framework, drawing on principles from self-supervised contrastive learning \cite{chen2020simple}. It combines unsupervised clustering with a multi-component loss function designed to enforce forgetting while preserving performance—all without using identity labels.

We use K-means clustering to generate dynamic pseudo-labels for the forget set, enabling cluster-based supervision. To unlearn effectively, we: (i) minimize cosine similarity between features from the unlearned and original models to encourage deviation, (ii) apply a pseudo-label loss to promote cluster reassignment, and (iii) introduce a centroid-guided contrastive loss that pushes forget features away from both the retain cluster centroids and their original features.
On the other hand, for the retain set, we maintain performance by: (i) maximizing cosine similarity with pre-unlearning representations, (ii) matching features with those from the original model, and (iii) applying feature distribution loss to preserve decision consistency.

\begin{figure*}[!t]
    \centering
    \includegraphics[width=\textwidth]{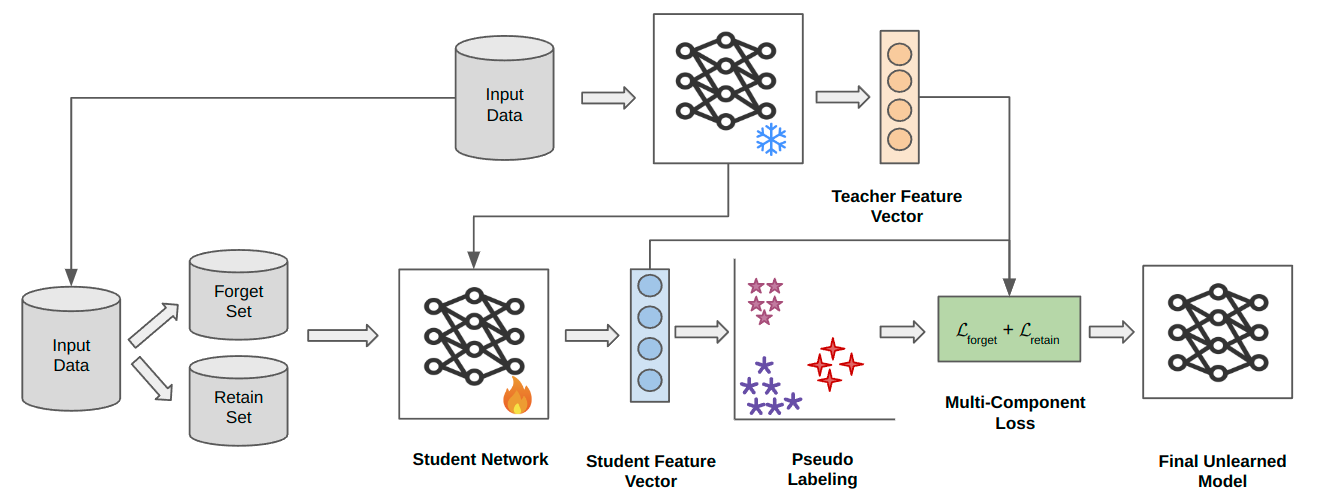}  
    \caption{Schematic diagram of the proposed CURE framework. Input data is split into forget (\( D_f \)) and retain (\( D_r \)). The student network, initialized from the teacher, processes them using pseudo-labeling for forget samples and multi-component loss (\( \mathcal{L}_{\text{forget}} + \mathcal{L}_{\text{retain}} \)) to balance forgetting and retention, giving the unlearned model.}
    \label{fig:framework}
\end{figure*}

In this study, we also identified a critical gap in how unlearning performance is measured. Existing metrics—such as forget set accuracy or retain set accuracy—fail to capture the trade-off between forgetting and retention. To address this, we introduce the \textbf{Unlearning Efficiency Score (UES)}, extending traditional evaluation frameworks to better reflect this balance in unlearning tasks. UES quantifies the drop in model performance on the forget set (indicating successful unlearning) while penalizing degradation on the retain set, offering a more holistic measure of unlearning quality that existing metrics overlook.


To evaluate the proposed CURE framework, we conduct experiments on the CASIA-WebFace dataset \cite{yi2014learning}, a widely used benchmark for facial recognition, with LFW \cite{huang2008labeled} and AgeDB \cite{moschoglou2017agedb} serving as verification datasets. We compare CURE against state-of-the-art unlearning methods, including unsupervised variants of BadTeacher, SCRUB, UNSIR \cite{tarun2023fast}, and the advanced Negative Gradient method \cite{choi2023towards}. CURE consistently outperforms all baselines across key metrics, demonstrating stronger forgetting, better retain set preservation, and improved generalization on LFW.
We also report results using our proposed UES metric, which provides a holistic measure of the trade-off between forgetting and retention.

Further, we also present a case study on unlearning low-quality images, where we partition the dataset based on image quality—assigning low-quality images to the forget set and high-quality images to the retain set. This setup enables us to evaluate performance under varying data quality conditions. This analysis reveals how verification accuracy varies with data quality, underscoring our method’s adaptability.



In summary, our contributions are as follows:
\begin{enumerate}

\item We propose the first unsupervised unlearning framework for face recognition, addressing a key gap by removing reliance on identity labels.
\item We present CURE, which combines unsupervised clustering with a multi-component loss to enforce forgetting while preserving performance—all without relying on identity labels.
\item We present UES as a principled metric that captures both forget set degradation and retain set preservation, addressing the gap left by prior evaluation methods.
\item We benchmark our approach against several unsupervised variants of state-of-the-art unlearning baselines 
and demonstrate superior performance in both accuracy drop and retain preservation.
\item We present a case study featuring a novel quality-aware analysis of the forget and retain sets.
\end{enumerate}

\section{Method}

\label{sec:methodology}

\subsection{Preliminary: Machine Unlearning}

In a standard\footnote{We use the term ``standard” loosely to refer to a typical supervised deep learning setup.}
 machine learning setup, a neural network \( f_{\theta} : \mathcal{X} \rightarrow \mathbb{R}^d \), parameterized by \( \theta \), is trained to map an input image \( x \in \mathcal{X} \) to a normalized \( d \)-dimensional embedding \( f_{\theta}(x) \in \mathbb{R}^d \). 
 Training is performed using a supervised objective function (e.g., cross-entropy loss) over a dataset \( D_{\text{train}} = \{(x_i, y_i)\}_{i=1}^N \), where \( y_i \) denotes the identity label for sample \( x_i \). After training, the model \( f_{\theta^*} \) is expected to achieve strong recognition performance, having encoded patterns from all training samples.

Machine unlearning modifies this standard setup by aiming to selectively erase the influence of a \textit{forget set} \( D_f \subset D_{\text{train}} \), while preserving the utility of the model on the \textit{retain set} \( D_r = D_{\text{train}} \setminus D_f \). The resulting model \( f_{\tilde{\theta}} \) should (i) produce representations for \( x_f \in D_f \) that significantly deviate from those produced by the original model \( f_{\theta}(x_f) \), effectively reducing the model’s ability to recall or rely on the forgotten data, (ii) preserve representations for \( x_r \in D_r \), such that \( f_{\tilde{\theta}}(x_r) \approx f_{\theta}(x_r) \), ensuring continuity in performance on retained data, and (iii) maintain generalization accuracy on unseen data, indicating that unlearning does not degrade the model’s broader capabilities.

The ideal benchmark for this task is \textit{Exact Unlearning}~\cite{cao2015towards}, which retrains a model from scratch on \( D_r \), ensuring full forgetting and retention. Practical approaches aim to approximate this behavior efficiently.

\subsection{CURE Framework}
In this section, we provide the details of our proposed \textit{(CURE)} framework.  
CURE is a machine unlearning framework that erases the influence of a forget set \( D_f \subset D_{\text{train}} \) from the trained model while retaining the features of a retain set \( D_r = D_{\text{train}} \setminus D_f \), without requiring identity labels.  
CURE uses a teacher-student framework~\cite{thudi2022unrolling}, where we use precomputed embeddings from a teacher model \( f_{\theta^*} \) to guide the student model \( f_{\tilde{\theta}} \) using novel strategies. 
The CURE pipeline involves the following steps: 
\begin{itemize} [leftmargin=*,noitemsep]
    \item \textbf{Precompute Teacher Embeddings}: We extract \( L_2 \)-normalized feature embeddings from the pretrained teacher model \( f_{\theta^*} \) for both \( D_f \) and \( D_r \).
    \item \textbf{Pseudo-Labeling of the Embeddings}: We apply K-means clustering to the normalized embeddings \( F_r \) and \( F_f \). For embeddings in \( F_f \), we assign pseudo-labels based on their farthest cluster centroid to encourage separation.
    \item \textbf{Optimize Student Model}: We train the student model \( f_{\tilde{\theta}} \) using a multi-component loss function guided by the pseudo-labels. During training, we periodically recompute forget cluster assignments using updated embeddings \( f_{\tilde{\theta}}(x_f) \) to adapt to the evolving feature space.
\end{itemize}
We present the schematic of CURE in Fig.~\ref{fig:framework}. We now describe each of these steps in detail.

\subsubsection{Teacher-Student Framework}
Our framework uses a teacher-student architecture for unlearning. The teacher network \( f_{\theta^*} : \mathcal{X} \rightarrow \mathbb{R}^d \), trained on \( D_{\text{train}} \), maps input images \( x \in \mathcal{X} \) to 
feature embeddings \( f_{\theta^*}(x) \in \mathbb{R}^d \).
We precompute these embeddings for the forget and retain sets as:
\begin{align}
F_f &= \{ f_{\theta^*}(x_f) \mid x_f \in D_f \}, \nonumber \\
F_r &= \{ f_{\theta^*}(x_r) \mid x_r \in D_r \}
\end{align}
The student network \( f_{\tilde{\theta}} \), initialized with \( \theta^* \) 
starts with the same configuration as the teacher and then undergoes the unlearning process to produce embeddings after removing the influence of the forget set \( D_{f} \) to produce unlearned embeddings \( f_{\tilde{\theta}}(x) \).

\subsubsection{Pseudo-Labeling with K-means Clustering}
To unlearn without relying on explicit labels, we propose a pseudo-labeling approach based on embedding clusters, enabling unsupervised unlearning. Specifically, we cluster the teacher’s forget embeddings \( F_f \) using the K-means clustering method~\cite{caron2020unsupervised,macqueen1967some}.
The forget centroids are represented as, \( C_f = \{ c_f^k \}_{k=1}^K \) and we assign each \( x_f \) a baseline label 
\[
l_f(x_f) = \arg\min_k \| f_{\theta^*}(x_f) - c_f^k \|_2.
\]
We then define a farthest cluster mapping~\cite{van2020scan} \( \phi : \{1, \ldots, K\} \rightarrow \{1, \ldots, K\} \) as:
\[
\phi(k) = \arg\max_{k'} \| c_f^k - c_f^{k'} \|_2,
\]
and assign pseudo-labels \( \tilde{y}_f(x_f) = \phi(l_f(x_f)) \). 
This reassignment aims to drive \( f_{\tilde{\theta}}(x_f) \) toward a new cluster, causing divergence from its original representation. This encourages the model to learn new decision boundaries for the forget samples that no longer resemble their original labels.
Forget clusters are dynamically updated every \( \tau \) epochs using \( f_{\tilde{\theta}} \) embeddings to adapt to the evolving feature space.
We also cluster the retain set embeddings using K-means and compute fixed centroids, denoted as \( C_r = \{ c_r^k \}_{k=1}^K \) from \( F_r \), which allows us to formulate an objective (described in Section~\ref{sec:loss_func}) that explicitly forces the forget set embeddings to stay away from the retain clusters, preventing them from drifting into the retain feature domain.

For these pseudo-labels, we use a linear classifier \( g : \mathbb{R}^d \to \mathbb{R}^K \), implemented as a fully connected layer, which maps the student’s feature embeddings \( f_{\tilde{\theta}}(x_f) \) to a probability distribution over the clusters. This classifier is trained jointly with \( f_{\tilde{\theta}} \) using the pseudo-label loss (described in Section~\ref{sec:loss_func}), encouraging the forget features to move toward a new target by aligning with their assigned farthest cluster.

\begin{figure*}[t]
    \centering
    \begin{subfigure}[t]{0.48\textwidth}
        \centering
        \includegraphics[width=\textwidth]{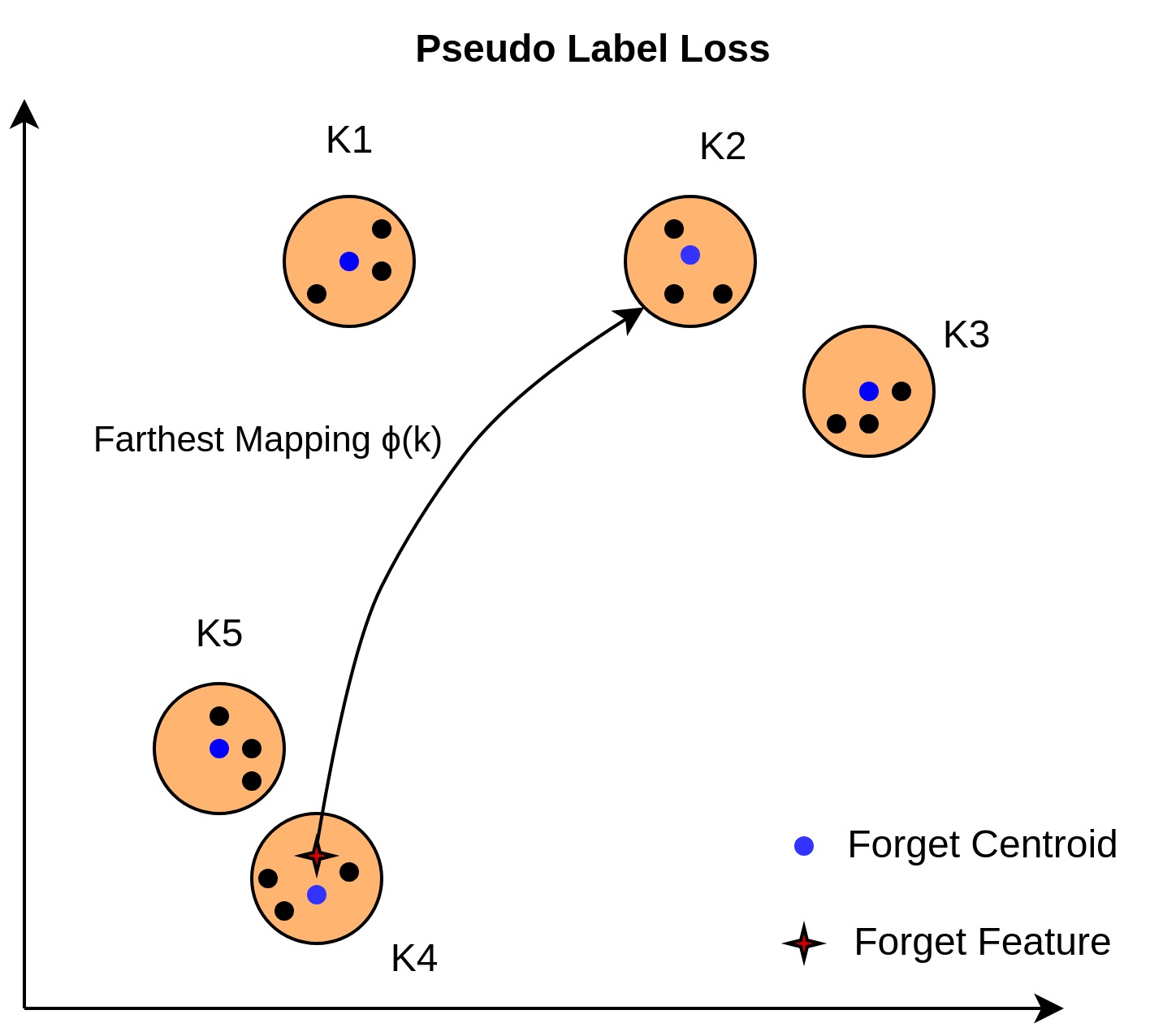}
        \caption{Diagram of the Pseudo-Labeling Loss, illustrating the drift of the student's forget feature \( f_{\tilde{\theta}}(x_f) \) from its initial cluster centroid towards the new distant assigned cluster via a linear classifier \( g \).}
        \label{fig:pseudo_label_loss}
    \end{subfigure}
    \hfill
    \begin{subfigure}[t]{0.48\textwidth}
        \centering
        \includegraphics[width=\textwidth]{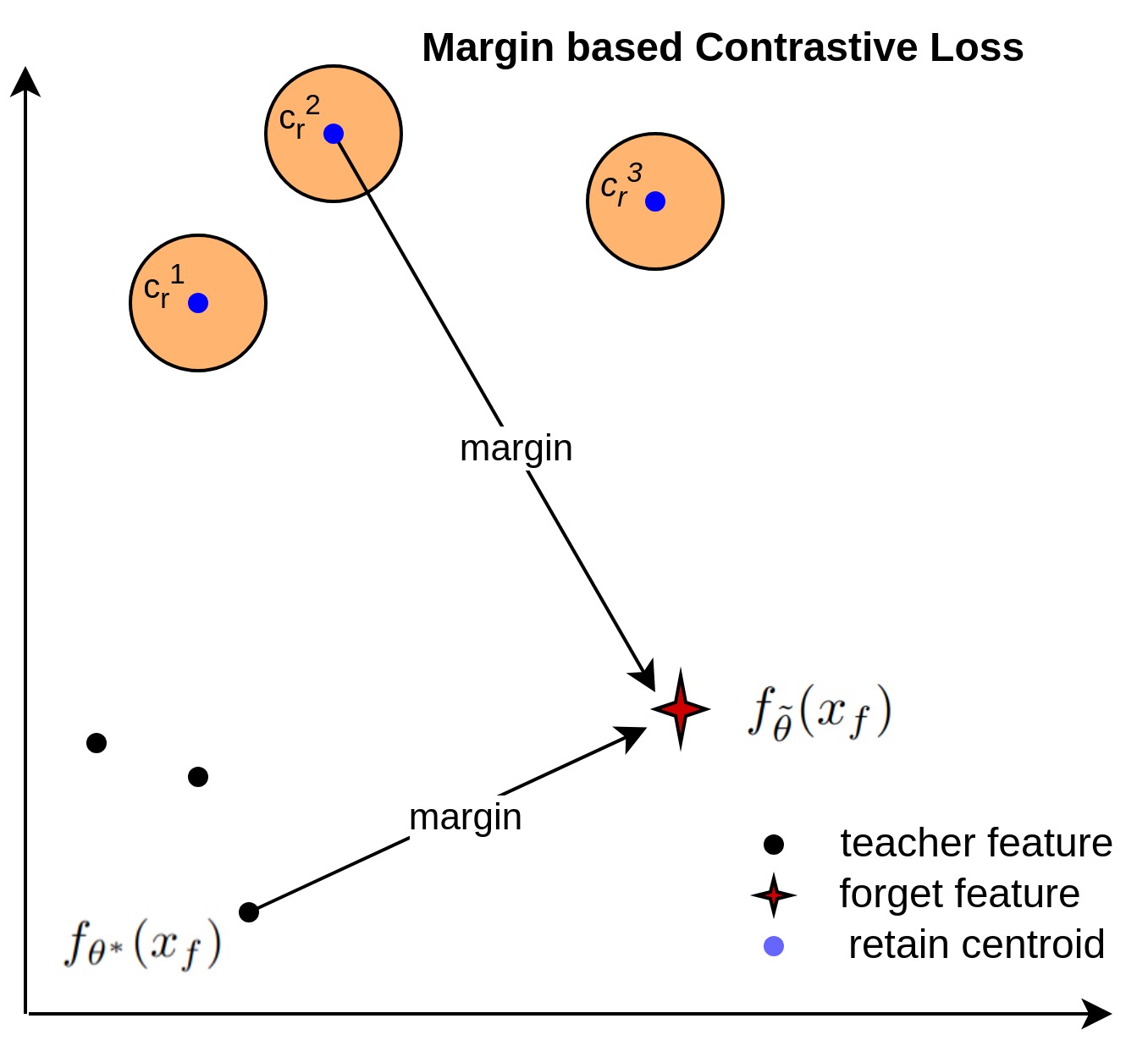}
        \caption{Illustration of the Contrastive Loss, enforcing separation of unlearned forget features \( f_{\tilde{\theta}}(x_f) \) from retain centroids \( c_r^k \) and their corresponding teacher features \( f_{\theta^*}(x_f) \) beyond a margin \( m = 0.3 \), assuming the constraints mentioned in the figure are satisfied.}
        \label{fig:contrastive_loss}
    \end{subfigure}
    \vspace{0.5cm}
    \caption{Illustrations of the two key forget loss components in our framework. }
\end{figure*}

\subsubsection{Loss Functions}
\label{sec:loss_func}
The proposed CURE framework optimizes the student model \( f_{\tilde{\theta}} \) and a linear classifier \( g : \mathbb{R}^d \rightarrow \mathbb{R}^K \) using a combination of forget and retain losses:
\[
\mathcal{L} = \mathcal{L}_{\text{forget}} + \mathcal{L}_{\text{retain}}.
\]
\paragraph{Forget Loss}
The forget loss \( \mathcal{L}_{\text{forget}} \) is carefully designed to erase the influence and any information of the forget set \( D_f \) from the student model using three complementary loss terms: a \textit{pseudo-label} loss, a \textit{cosine} loss, and a \textit{contrastive} loss. 
These terms together ensure that the student's forget features \( f_{\tilde{\theta}}(x_f) \) diverge from the teacher's original embeddings \( f_{\theta^*}(x_f) \) while remaining separated from the retain set \( D_r \).

\begin{itemize}[leftmargin=*,noitemsep]
\item \textbf{Pseudo-Label Loss}: This term guides the forget features toward new target based on farthest cluster assignments, extending pseudo-labeling techniques~\cite{lee2013pseudo,sohn2020fixmatch}:
    \begin{equation}
    \mathcal{L}_{\text{pl}} = \frac{1}{|D_f|} \sum_{x_f \in D_f} 
    \text{CE}\left( \frac{g(f_{\tilde{\theta}}(x_f))}{T}, \tilde{y}_f(x_f) \right),
    \end{equation}
    where \( \text{CE} \) denotes the cross-entropy loss, \( g \) is the classifier head, \( T \) is the temperature parameter, and \( \tilde{y}_f(x_f) \) represents the farthest cluster label assigned to each forget feature. By reassigning \( f_{\tilde{\theta}}(x_f) \) to the farthest cluster targets, this loss drives the forget features to drift toward new clusters, away from their original clusters, as schematically illustrated in Fig.~\ref{fig:pseudo_label_loss}, thereby enhancing forgetting efficiency.

    \item \textbf{Cosine Similarity Loss}: This term reduces the alignment between the student and teacher forget embeddings, encouraging the updated forget features to diverge from their original representations by minimizing their cosine similarity:
    \begin{equation}
    \mathcal{L}_{\text{cos}} = \frac{1}{|D_f|} \sum_{x_f \in D_f} f_{\tilde{\theta}}(x_f)^\top f_{\theta^*}(x_f),
    \end{equation}
    Since cosine similarity ranges from \(-1\) to \(1\), minimizing this term pushes the student’s features toward orthogonal or negatively aligned directions relative to their original embeddings, thereby disrupting their similarity.

    \item \textbf{Contrastive Loss}: To further enforce the structured forgetting and separation between forget features and retain features, we use a margin‑based contrastive loss defined as:
  \begin{equation}
    \label{eq:cont_loss}
    \begin{split}
      \mathcal{L}_{\mathrm{cont}}
      &= \frac{1}{|D_f|}\sum_{x_f\in D_f}\Bigl[\, 
         \mathrm{ReLU}\bigl(f_{\tilde{\theta}}(x_f)^\top f_{\theta^*}(x_f)
          \\
      &\quad\;
      - m\bigr) +\
         \mathrm{ReLU}\bigl(f_{\tilde{\theta}}(x_f)^\top 
          c_r^{l_f(x_f)} - m\bigr)\Bigr]
    \end{split}
  \end{equation}
    Here, \(m\) is the similarity margin, and \(c_r^{l_f(x_f)}\) represents the retain-cluster centroid. The two terms are complementary: the first penalizes forget features for being close to the teacher embeddings, while the second penalizes them for being close to the retain set features.
    We present the schematic diagram of this loss in Fig.~\ref{fig:contrastive_loss}.  

\end{itemize}   
  
The total forget loss is:
\begin{equation}
\begin{split}
\mathcal{L}_{\text{forget}} =  \mathcal{L}_{\text{pl}} 
+  \mathcal{L}_{\text{cos-forget}}
+  \mathcal{L}_{\text{contrast}},
\end{split}
\end{equation}

\paragraph{Retain Loss}
The retain loss \( \mathcal{L}_{\text{retain}} \) is carefully designed to preserve the retain features and their information intact in the student model. We use a \textit{cosine} similarity loss, a \textit{feature matching} loss, and a \textit{feature distribution} loss. These components are complementary and collectively ensure that the student’s retain features \( f_{\tilde{\theta}}(x_r) \) closely align with the teacher’s original retain features \( f_{\theta^*}(x_r) \), maintaining retain accuracy and generalization.
\begin{itemize} [leftmargin=*,noitemsep]
    \item \textbf{Cosine Similarity Loss}: This term maximizes the alignment between between student and teacher retain embeddings:
    \[
    \mathcal{L}_{\text{cos-retain}} = -\frac{1}{|D_r|} \sum_{x_r \in D_r} f_{\tilde{\theta}}(x_r)^\top f_{\theta^*}(x_r),
    \]
    By minimizing the negative cosine similarity, we encourage \( f_{\tilde{\theta}}(x_r) \) to align closely with \( f_{\theta^*}(x_r) \) in direction, thereby preserving the angular consistency essential for retaining \( D_r \).

    \item \textbf{Feature Matching Loss}: This term minimizes the squared Euclidean distance between features:
    \[
    \mathcal{L}_{\text{feat}} = \frac{1}{|D_r|} \sum_{x_r \in D_r} \| f_{\tilde{\theta}}(x_r) - f_{\theta^*}(x_r) \|_2^2,
    \]
    It ensures fine-grained preservation of the retain embeddings by encouraging precise pointwise preservation. 

    \item \textbf{Feature Distribution Loss}: This encourages the student’s retain feature embeddings to match the teacher’s retain distributional properties:
    \begin{equation}
    \begin{split}
    \mathcal{L}_{\text{fd}} = \frac{T^2}{|D_r|} \sum_{x_r \in D_r} \text{KL}\Big( 
    \text{softmax}\Big(\frac{f_{\tilde{\theta}}(x_r)}{T}\Big), \\
    \text{softmax}\Big(\frac{f_{\theta^*}(x_r)}{T}\Big) \Big),
    \end{split}
    \end{equation}
    where \( \text{KL} \) is the Kullback-Leibler divergence, and T scales the softmax applied to the \( L_2 \)-normalized feature vectors. This term enforces distribution-level alignment across the feature dimensions, further enhancing generalization to unseen datasets.
\end{itemize}
Overall, the total retain loss is:
\begin{equation}
\begin{split}
\mathcal{L}_{\text{retain}} = \mathcal{L}_{\text{fd}} +  \mathcal{L}_{\text{feat}} +  \mathcal{L}_{\text{cos-retain}}
\end{split}
\end{equation}

\textbf{Design Rationale}: 
The proposed combination of the loss terms for the forget and retain sets are chosen to balance forgetting and retention efficiency.
For the \textit{forget set}, we designed a structured forgetting strategy inspired by representation learning techniques, to explicitly enforce separation in the feature space. Minimizing cosine similarity enforces directional separation, using pseudo labeling reassigns forget features to new clusters and then using contrastive loss enforces separation between both the original identity and nearby retain centroids. 
For the \textit{retain set}, feature matching and cosine similarity losses jointly preserve angular alignment, magnitude, and complementary gradient behaviors, ensuring both pointwise and directional structure preservation. The feature distribution loss further captures global structural and distributional alignment in the feature space. We also emphasize that this loss formulation is validated both empirically and through ablation analysis (Section \ref{sec:result}).

The overall algorithm for the CURE framework is presented in Algorithm \ref{alg:cure}.

\begin{algorithm}[H]
\caption{ CURE (Centroid-guided Unsupervised Representation Erasure)}
\label{alg:cure}
\begin{algorithmic}[1]
\Require Pretrained model $f_{\theta^*}$, forget set $D_f$, retain set $D_r$, clusters $K$, margin $m$, temperature $T$, epochs $E$, update interval $\tau$
\Ensure Unlearned model $f_{\tilde{\theta}}$
\State Initialize $f_{\tilde{\theta}} \gets f_{\theta^*}$; freeze early layers
\State Compute teacher features: $F_f \gets f_{\theta^*}(D_f)$, $F_r \gets f_{\theta^*}(D_r)$
\State K-means on $F_r$: fixed retain centroids $C_r = \{c_r^k\}_{k=1}^{K}$
\State K-means on $F_f$: initial forget centroids $C_f = \{c_f^k\}_{k=1}^{K}$
\State For each cluster $k$, define farthest cluster mapping: $\phi(k) = \arg\max_{k'} \|c_f^k - c_f^{k'}\|_2$
\For{$e = 1$ to $E$}
    \State Sample batches $B_f \subset D_f$, $B_r \subset D_r$
    \State Compute student embeddings: $f_{\tilde{\theta}}(B_f)$, $f_{\tilde{\theta}}(B_r)$
    \State Assign pseudo-labels to $x_f \in B_f$: 
    \Statex \hspace{\algorithmicindent} $l_f(x_f) = \arg\min_k \|f_{\theta^*}(x_f) - c_f^k\|_2$
    \Statex \hspace{\algorithmicindent} $\tilde{y}_f(x_f) = \phi(l_f(x_f))$
    \State Compute total loss:
    \Statex \hspace{\algorithmicindent} $\mathcal{L}_{\text{forget}} = \mathcal{L}_{\text{pl}} + \mathcal{L}_{\text{cos-forget}} + \mathcal{L}_{\text{contrast}}$
    \Statex \hspace{\algorithmicindent} $\mathcal{L}_{\text{retain}} = \mathcal{L}_{\text{cos-retain}} + \mathcal{L}_{\text{feat}} + \mathcal{L}_{\text{fd}}$
    \Statex \hspace{\algorithmicindent} $\mathcal{L} = \mathcal{L}_{\text{forget}} + \mathcal{L}_{\text{retain}}$
    \State Update $f_{\tilde{\theta}}$ via SGD using $\mathcal{L}$
    \If{$e \bmod \tau = 0$}
        \State Recompute $C_f$ using K-means on $f_{\tilde{\theta}}(D_f)$
        \State Update $\phi(k)$ accordingly
    \EndIf
\EndFor
\State \Return $f_{\tilde{\theta}}$
\end{algorithmic}
\end{algorithm}

\subsection{UES: MU evaluation metric}
\label{sec:evaluation}
In machine unlearning, unlearning 
is quantified by the model's ability to forget 
\( D_f \) while retaining \( D_r \).
However, most existing evaluation metrics 
struggle to capture both objectives simultaneously, often being 
biased towards forgetting or retention~\cite{bourtoule2021machine, sekhari2021remember, thudi2022necessity, golatkar2020eternal, golatkar2021mixed, golatkar2020forgetting, graves2021amnesiac}. For instance, metrics such as membership inference attacks~\cite{sekhari2021remember, carlini2021extracting} neglect retention by focusing on forgetting and metrics such as, confidence drop~\cite{kurmanji2023towards} and entropy increase~\cite{yeom2018privacy} are either biased towards forgetting or retention.
To address this, we introduce the \textbf{Unlearning Efficiency Score (UES)} as a unified metric that jointly measures both aspects. 
UES is defined as:
\begin{small}
\begin{multline}
\text{UES} = \alpha \cdot \text{Normalized Forget Drop} \\
 - (1 - \alpha) \cdot \text{Normalized Retain Drop}
\end{multline}
\end{small}

\begin{align*}
\text{Normalized Forget Drop} &= \frac{\text{Acc}_{\theta^*}(D_f) - \text{Acc}_{\tilde{\theta}}(D_f)}{\text{Acc}_{\theta^*}(D_f)}, \\
\text{Normalized Retain Drop} &= \frac{\text{Acc}_{\theta^*}(D_r) - \text{Acc}_{\tilde{\theta}}(D_r)}{\text{Acc}_{\theta^*}(D_r)}.
\end{align*}
Here, \( \text{Acc}_{\theta^*}(D_f) \) and \( \text{Acc}_{\theta^*}(D_r) \) denote the accuracies on the forget and retain sets before unlearning, while \( \text{Acc}_{\tilde{\theta}}(D_f) \) and \( \text{Acc}_{\tilde{\theta}}(D_r) \) are their accuracies after unlearning. The use of parameter \( \alpha \in [0, 1] \) balances these objectives.


We set \( \alpha = 0.5 \) to balance forgetting and retention, aligning with our goal of achieving a fair trade-off for facial recognition unlearning systems. However, depending on the application, the UES and the choice of \( \alpha \) offer \textit{flexibility} to prioritize either aspect. Specifically, setting \( \alpha < 0.5 \) emphasizes retention over forgetting, while \( \alpha > 0.5 \) places greater emphasis on forgetting over retention.
\newline
\textbf{Other evaluation metrics:}
In addition to UES, the efficiency of the proposed unlearning method is evaluated by comparing classification performance before and after unlearning using complementary metrics inspired by prior works~\cite{kurmanji2023towards, tarun2023fast, gundougdu2024study, golatkar2021mixed}. 

We further assess shifts in model uncertainty using confidence and entropy:
\begin{align}
\text{Conf} &= \frac{1}{N} \sum_{i=1}^N \max_j P(y_j | x_i), \\
H &= -\frac{1}{N} \sum_{i=1}^N \sum_{j=1}^C P(y_j | x_i) \log P(y_j | x_i).
\end{align}

In addition, we evaluate membership inference~\cite{sekhari2021remember, carlini2021extracting} characteristics on forget dataset by calculating:
\begin{itemize} [leftmargin=*,noitemsep]
    \item \textit{Activation Distance}: L2 distance between softmax outputs before and after unlearning.
    \item \textit{Layer-wise Distance}: Average L2 norm between model parameters.
    \item \textit{Completeness}: Jaccard similarity of top-3 predictions.
    \item \textit{Membership Recall}: Fraction of forget samples with prediction confidence \( > 0.8 \).
\end{itemize}
Together, these complementary metrics offer deeper insights into the unlearning model’s forgetting effectiveness and retention stability.


\section{Experiments}
\label{sec:experiments}

\subsection{Dataset and Preprocessing}
\label{subsec:dataset}

CASIA-WebFace \cite{yi2014learning} was used for training our model—both the teacher network and later our proposed CURE framework. The dataset contains approximately 490{,}623 images across 10{,}572 unique classes. For the unlearning setup, we randomly partitioned this dataset into a forget set \( D_f \), comprising 1{,}222 classes and 36{,}393 images, and a retain set \( D_r \), consisting of 9{,}350 classes and 454{,}230 images. 
To ensure consistency and image quality, all raw CASIA-WebFace images were preprocessed using the MTCNN framework~\cite{zhang2016joint} for face detection and alignment. The resulting images were resized to a standard resolution of \( 112 \times 112 \) pixels. 
For validation, we used LFW and AgeDB dataset. 
These datasets were preprocessed using the same pipeline as CASIA-WebFace. 


\subsection{Training}

The face recognition model is trained using a Res100-IR backbone architecture~\cite{he2016deep} with ArcFace~\cite{deng2019arcface} as the margin-based loss layer. The model is optimized using
Adam optimizer 
with an initial learning rate of 0.04, momentum of 0.9, and weight decay of \( 5 \times 10^{-4} \), over the course of 50 epochs. We employ a multi-step learning rate scheduler with decay milestones at epochs 6, 11, and 16, using a decay factor (gamma) of 0.1. The training is conducted with a batch size of 60. Standard data augmentation techniques are applied, including normalization.


For CURE, we adopt the same optimizer configuration, using an initial learning rate of \( 10^{-4} \), momentum of 0.9, and weight decay of \( 5 \times 10^{-4} \), applied over 50 epochs with a batch size of 128. The learning rate decays at epochs 5 and 10 with a decay factor (gamma) of 0.1. 
To reflect the evolving representation \( f_{\tilde{\theta}}(x_f) \), forget clusters are recomputed every 5 epochs. We use \( K = 100 \) clusters for pseudo-labeling, a contrastive margin of \( m = 0.3 \), and a temperature parameter \( T = 0.5 \) to soften the probability distributions. 
The loss weights are configured such that \( \lambda_{\text{fd}} = 10 \), while all other loss terms are assigned a weight of 1.


\section{Results}
\label{sec:result}

We evaluate CURE against several existing machine unlearning approaches, including SCRUB~\cite{kurmanji2023towards}, BadTeacher~\cite{chundawat2023can}, UNSIR~\cite{tarun2023fast}, and Adv NegGrad~\cite{graves2021amnesiac}. Our goal is to assess how CURE performs relative to these methods across multiple metrics: accuracy on the forget set (Forget Acc), accuracy on the retain set (Retain Acc), the balance between these objectives as captured by our proposed UES, and overall generalization performance evaluated via verification.
The results are summarized in Table~\ref{tab:unlearning_results}. In terms of forgetting, CURE ranks second to SCRUB; however, SCRUB demonstrates significantly lower retain accuracy. Conversely, while CURE's retain accuracy is comparable to Adv NegGrad and lower than BadTeacher and UNSIR, those methods exhibit poor forget accuracy. Overall, CURE achieves the best balance between forgetting and retention, as reflected in the highest UES score.
For external verification, CURE also achieves competitive performance compared to other methods, further demonstrating its generalization ability.

\begin{table*}[!t]
\centering
\small
\caption{Comparison of unlearning methods across (\(D_f\)), (\(D_r\)), and generalization on unseen datasets (LFW, AgeDB). 
Before unlearning represents the accuracies before unlearning.}
\vspace{0.5cm}
\label{tab:unlearning_results}
\begin{tabular}{l c c c cc}
\toprule
\multirow{2}{*}{\textbf{Method}} & 
\multirow{2}{*}{\textbf{Forget Acc (\%)~$\downarrow$}} & 
\multirow{2}{*}{\textbf{Retain Acc (\%)~$\uparrow$}} & 
\multirow{2}{*}{\textbf{UES~$\uparrow$}} & 
\multicolumn{2}{c}{\textbf{Balanced Verification}} \\
\cmidrule(lr){5-6}
& & & & \textbf{LFW (\%)} & \textbf{AgeDB (\%)} \\
\midrule
Before Unlearning & 80.85 & 87.12 & - & 98.03 & 94.33 \\
\hline
SCRUB & 6.60 & 19.08 & 0.0687 & 96.03 & 91.95 \\
BadTeacher & 30.09 & 67.17 & 0.1994 & 96.30 & 91.65 \\
UNSIR & 20.74 & 60.17 & 0.2171 & 97.48 & 94.63 \\
Adv NegGrd & 20.78 & 45.01 & 0.1298 & 98.05 & 94.43 \\
\textbf{CURE (Ours)} & \textbf{6.91} & \textbf{47.34} & \textbf{0.2290} & \textbf{97.05} & \textbf{92.85} \\
\bottomrule
\end{tabular}
\end{table*}


\begin{table*}[!t]
\centering
\small
\caption{Evaluation of Confidence Drop, Entropy Increase, and Membership Inference Metrics. Membership recall before unlearning is 0.4463, Mem Recall in table shows after unlearning results. }
\vspace{0.5cm}
\label{tab:combined_metrics}
\begin{tabular}{l c c c c c c c c}
\toprule
\multirow{2}{*}{\textbf{Method}} & 
\multicolumn{2}{c}{\textbf{Forget (\( D_f \))~$\uparrow$}} & 
\multicolumn{2}{c}{\textbf{Retain (\( D_r \))~$\downarrow$}} & 
\multicolumn{4}{c}{\textbf{Membership Inference Results}} \\
\cmidrule(lr){2-3} \cmidrule(lr){4-5} \cmidrule(lr){6-9}
& Conf Drop & Ent Inc & Conf Drop & Ent Inc & Act Dis~$\uparrow$ & Layer Dis~$\uparrow$ & Comp~$\downarrow$ & Mem Recall~$\downarrow$ \\
\midrule
SCRUB & 0.4614 & 2.3948 & 0.4850 & 2.5969 & 0.6900 & 114918.1738 & 0.1291 & 0.0236 \\
BadTeacher & 0.4656 & 2.6407 & 0.0543 & 0.3306 & 0.5068 & 22341.9058 & 0.3649 & 0.0692 \\
UNSIR & 0.3662 & 1.5755 & 0.0210 & 0.0430 & 0.6370 & 88715.7954 & 0.2143 & 0.0904 \\
Adv NegGrd & 0.4556 & 2.2511 & 0.2760 & 1.7580 & 0.5982 & 117872.5402 & 0.2373 & 0.0253 \\
\textbf{CURE (Ours)} & \textbf{0.5057} & \textbf{2.6298} & \textbf{0.0808} & \textbf{0.3912} & \textbf{0.6663} & \textbf{147340.5812} & \textbf{0.1809} & \textbf{0.0132} \\
\bottomrule
\end{tabular}
\end{table*}

To further evaluate the quality of representations obtained from CURE compared to other unlearning methods, we assess the confidence drop and entropy increase on both the forget and retain sets, and additionally analyze the membership inference results (as discussed earlier in Section~\ref{sec:evaluation}). These results are presented in Table~\ref{tab:combined_metrics}.
As shown, CURE achieves the highest confidence drop (0.5057) and a substantial entropy increase (2.6298) on \( D_f \), indicating strong uncertainty injection into the forgotten predictions. On \( D_r \), CURE maintains superior performance with a minimal confidence drop (0.0808) and a small entropy increase (0.3912), preserving model stability and retention capabilities.
The membership inference results on \( D_f \) further highlight CURE's privacy benefits: its post-unlearning membership recall is 0.0132, the lowest among all evaluated methods, suggesting effective obfuscation of forgotten samples. CURE also achieves the best layer-wise distance along with a high activation distance, demonstrating strong forgetting. Additionally, it exhibits lower completeness, indicating successful removal of the targeted samples.



\subsection{Face Image Quality and Unlearning}

\begin{figure}[!t]
    \centering
    \includegraphics[width=\columnwidth]{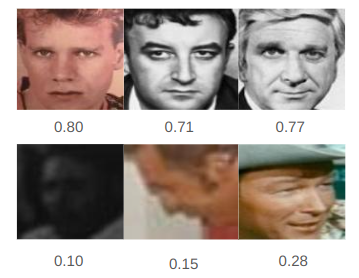}
    \caption{High-quality images (top row) and low-quality images (bottom row) from the AgeDB dataset, along with their corresponding quality scores based on the FIQA metric.}
    \label{fig:img3Example}
\end{figure}

This section presents a practical scenario where face image quality defines the unlearning criterion. With legacy data and models being reused across various biometric systems, unlearning offers a viable approach to refine existing models by removing the influence of poor-quality data. 
To demonstrate this use case, we first utilize FIQA, a face image quality assessment method~\cite{najafzadeh2023face}, which provides a quality score ranging from 0 to 1. We compute quality scores for all images in our dataset. Examples of high-quality and low-quality images, along with their corresponding scores, are shown in Fig.~\ref{fig:img3Example}.
We categorize the images into low-quality (LQ, \(n = 171{,}717\)) and high-quality (HQ, \(n = 318{,}906\)) groups using the 35th percentile as a threshold. LQ images are designated as the forget set, while HQ images form the retain set.
We then evaluate external verification accuracy across three different settings of the AgeDB dataset: the full AgeDB, a low-quality AgeDB subset, and a high-quality AgeDB subset. The latter two are partitioned based on image quality using the same procedure described above.


\begin{table}[t]
\centering
\footnotesize
\setlength{\tabcolsep}{5pt}
\caption{Verification Accuracy after unlearning on AgeDB, LQ AgeDB, and HQ AgeD.}
\label{tab:verification_metrics}
\begin{tabular}{lccc}
\toprule
\textbf{Dataset} & \textbf{AgeDB} & \textbf{LQ AgeDB} & \textbf{HQ AgeDB} \\
\midrule
Before unleanring   & 94.33 & 94.00 & 95.33 \\
\hline
CURE - random forget   & 92.85 & 92.38 & 93.28 \\
CURE - LQ forget    & 94.50 & 93.75 & 94.90 \\
\bottomrule
\end{tabular}
\end{table}

We present the results in Table~\ref{tab:verification_metrics}. As shown, forgetting low-quality images from the training data leads to an improvement in overall AgeDB accuracy compared to the baseline model (i.e., before unlearning). 
The accuracy on the LQ subset of AgeDB drops slightly, indicating effective forgetting of low-quality samples, while the HQ subset shows only a marginal decline, suggesting that the model retains its performance on high-quality data. 
It is important to note that all previous analyses were based on randomly selected forget sets (CURE-random forget). In contrast, this case study—CURE-LQ Forget—demonstrates a clear advantage over such random forgetting. 
We believe this is the first study to explore the role of image quality in machine unlearning, and it offers a promising direction for future research on quality-aware unlearning.

\section{Ablation Study}
\label{sec:ablation}

In this section, we present an ablation analysis to evaluate key design choices in our study. Specifically, we examine the contributions of the multi-component retain and forget loss terms, the use of \(K\)-means clustering for pseudo-labeling, and the impact of varying margin values in the contrastive loss. The overall results are summarized in Table~\ref{tab:ablation}.
We begin with our proposed configuration (CURE), followed by a series of ablations: removal of retain loss components (Ablations 1--6), removal of forget loss components (Ablations 7--12), modifications to the pseudo-labeling strategy (Ablations 13--14), and variations in margin values (Ablations 15--17).



\begin{table}[t]
\centering
\caption{Ablation results showing Forget accuracy after unlearning, Retain accuracy after unlearning), and the UES. For all ablations, the Forget accuracy before unlearning is 80.85\% and the Retain accuracy before unlearning is 87.12\%.}
\vspace{0.5cm}
\label{tab:ablation}
\footnotesize
\setlength{\tabcolsep}{5pt}
\begin{tabular}{lccc}
\toprule
\textbf{Ablation} & \textbf{Forget} & \textbf{Retain} & \textbf{UES} \\
\midrule
\textbf{CURE} & \textbf{6.91} & \textbf{47.34} & \textbf{0.229} \\
\midrule
\multicolumn{4}{c}{\emph{Retain Loss Ablations}} \\
A1: \( \mathcal{L}_{\text{cos-retain}} \) & 14.09 & 34.21 & 0.109 \\
A2: \( \mathcal{L}_{\text{feat}} \) & 0.00 & 0.00 & 0.000 \\
A3: \( \mathcal{L}_{\text{fd}} \) & 0.00 & 16.56 & 0.095 \\
A4: \( \mathcal{L}_{\text{feat}} + \mathcal{L}_{\text{cos}} \) & 19.15 & 39.98 & 0.111 \\
A5: \( \mathcal{L}_{\text{fd}} + \mathcal{L}_{\text{feat}} \) & 0.00 & 17.30 & 0.099 \\
A6: \( \mathcal{L}_{\text{cos}} + \mathcal{L}_{\text{fd}} \) & 3.94 & 41.70 & 0.215 \\
\midrule
\multicolumn{4}{c}{\emph{Forget Loss Ablations}} \\
A7: \( \mathcal{L}_{\text{pl}} \) & 73.42 & 84.60 & 0.031 \\
A8: \( \mathcal{L}_{\text{cont}} \) & 70.29 & 83.83 & 0.046 \\
A9: \( \mathcal{L}_{\text{cos-forget}} \) & 1.28 & 35.28 & 0.195 \\
A10: \( \mathcal{L}_{\text{cos-forget}} + \mathcal{L}_{\text{pl}} \) & 0.34 & 31.65 & 0.180 \\
A11: \( \mathcal{L}_{\text{cos-forget}} + \mathcal{L}_{\text{cont}} \) & 9.00 & 48.29 & 0.222 \\
A12: \( \mathcal{L}_{\text{cont}} + \mathcal{L}_{\text{pl}} \) & 30.11 & 68.83 & 0.209 \\
\midrule
\multicolumn{4}{c}{\emph{Pseudo Labeling based Ablations}} \\
A13: {\scriptsize CURE-GMM} & 9.96 & 48.87 & 0.2189 \\
A14: {\scriptsize CURE-DBSCAN} & 2.63 & 27.72 & 0.1428 \\
\midrule
\multicolumn{4}{c}{\emph{Margin Ablations}} \\
A15: \( m = 0 \) & 4.07 & 41.28 & 0.212 \\
A16: \( m = 0.1 \) & 6.43 & 44.72 & 0.217 \\
A17: \( m = 0.6 \) & 4.99 & 43.71 & 0.220 \\
\bottomrule
\end{tabular}
\end{table}

As shown in Table~\ref{tab:ablation}, removing any component from \( \mathcal{L}_{\text{forget}} \) or \( \mathcal{L}_{\text{retain}} \) significantly degrades the unlearning model's performance, underscoring the importance of our full multi-component loss design.
Within the retain loss, \(\mathcal{L}_{\text{fd}}\) and \( \mathcal{L}_{\text{cos-retain}} \) are particularly critical. However, when used in isolation, individual components—especially \(\mathcal{L}_{\text{feat}}\)—are not sufficient to ensure strong performance. Notably, combining \(\mathcal{L}_{\text{feat}}\) with \(\mathcal{L}_{\text{fd}}\) and \( \mathcal{L}_{\text{cos-retain}} \) enhances both retention and overall unlearning effectiveness by preserving the magnitude of student-teacher embedding distances and anchoring fine-grained spatial alignment.
For the forget loss, omitting \(\mathcal{L}_{\text{cos-forget}}\) leads to a clear drop in forgetting, but when it is combined with \(\mathcal{L}_{\text{pl}}\) and \(\mathcal{L}_{\text{cont}}\), the model achieves improved forgetting 
with competitive retention performance.

In our pseudo-labeling ablation, we evaluated the CURE framework with two alternative clustering algorithms: Density-Based Spatial Clustering of Applications with Noise (DBSCAN) and Gaussian Mixture Models (GMM). While each has its strengths and limitations depending on context~\cite{kanagala2016comparative, murugesan2019benchmarking}, K-means consistently performed best in our setup, mainly due to ArcFace feature embeddings being naturally clustered from discriminative training. Finally, we analyzed the effect of different margin values in the contrastive loss and observed that a margin of \( m = 0.3 \) yields the best performance. However, the overall variation across different margins was not substantial.

\section{Conclusion}

In this paper, we presented CURE, a novel unsupervised unlearning method designed for facial recognition systems that operates without identity labels. It leverages a student-teacher paradigm and applies dedicated forget and retain loss functions to selectively erase the influence of the forget set while preserving the effect of the retain set.
We also introduced the Unlearning Efficiency Score (UES), a new metric that jointly quantifies balanced forgetting and retention.
Across all experiments, CURE demonstrated competitive or superior performance compared to several existing methods, particularly in terms of UES. We also conducted the first case study on the role of image quality in unlearning, showing that forgetting low-quality images can improve performance beyond random forgetting and even achieve baseline performance.
Future research may explore quality-aware and task-adaptive unlearning strategies, as well as the development of scalable and generalizable unlearning frameworks applicable to other biometric modalities beyond facial recognition.

\section{Acknowledgments}
This research is supported by the Center for Identification Technology Research and the National Science Foundation under Grant No. 1650474.

{\small
\bibliographystyle{ieee}
\bibliography{egbib}
}



\end{document}